\documentclass[a4paper,conference,10pt]{IEEEtran}
\usepackage{graphicx}
\usepackage[backend=bibtex]{biblatex} 
\usepackage{multirow}
\usepackage[tableposition=top]{caption}
\usepackage{xcolor}
\usepackage[many]{tcolorbox}
\usepackage{amsmath}
\usepackage{amssymb, subcaption, caption}
\begin{document}

\title{Metamorphic Testing-based Adversarial Attack to Fool Deepfake Detectors}
\makeatletter
\newcommand{\linebreakand}{%
  \end{@IEEEauthorhalign}
  \hfill\mbox{}\par
  \mbox{}\hfill\begin{@IEEEauthorhalign}
}

\author{
\IEEEauthorblockN{Nyee Thoang Lim\IEEEauthorrefmark{1}, Meng Yi Kuan\IEEEauthorrefmark{1}, Muxin Pu\IEEEauthorrefmark{1}, Mei Kuan Lim\IEEEauthorrefmark{1}, Chun Yong Chong\IEEEauthorrefmark{1}}
\IEEEauthorblockA{\IEEEauthorrefmark{1}School of Information Technology,\\ 
Monash University Malaysia, 47500, Bandar Sunway, Selangor, Malaysia\\
\{nlim0019, mkua0001, mpuu0001\}@student.monash.edu  \{lim.meikuan, chong.chunyong\}@monash.edu}
}

\maketitle

\begin{abstract}
Deepfakes utilise Artificial Intelligence (AI) techniques to create synthetic media where the likeness of one person is replaced with another. There are growing concerns that deepfakes can be maliciously used to create misleading and harmful digital contents. As deepfakes become more common, there is a dire need for deepfake detection technology to help spot deepfake media. Present deepfake detection models are able to achieve outstanding accuracy ($>$90\%). However, most of them are limited to within-dataset scenario. Most models do not generalise well enough in cross-dataset scenario. Furthermore, state-of-the-art deepfake detection models rely on neural network-based classification models that are known to be vulnerable to adversarial attacks. Motivated by the need for a robust deepfake detection model, this study adapts metamorphic testing (MT) principles to help identify potential factors that could influence the robustness of the examined model, while overcoming the test oracle problem in this domain. Metamorphic testing is specifically chosen as the testing technique as it fits our demand to address learning-based system testing with probabilistic outcomes from largely black-box components, based on potentially large input domains. We performed our evaluations on MesoInception-4 and TwoStreamNet models, which are the state-of-the-art deepfake detection models. This study identified makeup application as an adversarial attack that could fool deepfake detectors. Our experimental results demonstrate that both the MesoInception-4 and TwoStreamNet models degrade in their performance by up to 30\% when the input data is perturbed with makeup.

\end{abstract}
\IEEEpeerreviewmaketitle

\section{Introduction}
Synthetic media are commonly known as deepfakes, which are hyper-realistic videos or images that use AI and deep learning to portray someone saying or doing things that never took place \cite{deepfakeMeaning}. With the advancement of technology, deepfakes raise growing public concerns about their potential to be used to spread misinformation and commit crimes. 


Simultaneously, the research effort in digital image forensics to detect image forgeries has grown to regulate the circulation of such falsified content \cite{8630761}. Current deepfake detection models adopt machine learning (ML) and deep learning (DL), which are two of the main approaches in AI. These models have shown excellent results in identifying deepfakes images or videos \cite{Shad2021}. 

Despite the glory of reported success cases, it is undeniable that AI models can often fail in unintuitive ways. One of the challenges that modern AI models face is domain adaptation \cite{zhou_2022}. As AI models are trained on the specific domain collected, their performance can be unsatisfactory on unseen domains due to the domains' distribution discrepancies, or bias \cite{zhou_2022}. Besides, the purposeful invisible modification of even a single pixel of the images can also easily induce the model to perform classification mistakes \cite{one_pixel_attack}. Thus, to fully adopt a trustworthy AI technology in real-world scenarios, proving its robustness through testing is an ongoing challenge due to its learning-based property. Although AI models' black-box nature is extremely powerful, the non-transparency brought along can cause practical problems which make them unreliable.

Researchers have shown various examples to fool neural network algorithms. The most prominent one is the work of a group of researchers at Google and New York University (NYU), which found that it is easy to fool Convolutional Neural Networks (CNNs) by adding carefully constructed noise to the input \cite{goodfellow2015explaining}. In another example, a study has shown that by maliciously adding noisy pixels to medical scans, it could fool a neural network-based model into wrongly detecting cancer \cite{finlayson2019adversarial}.

However, there is few evidence to show whether anyone can fool deepfake detection models by adding carefully crafted noise to the input dataset, which we often refer to as an adversarial attack on a neural network. Adversarial attacks on deepfake detectors can be detrimental. The potential of sabotaging deepfake detectors simply by adding some artifacts or noise to bypass detection can easily fool these models to wrongly classify deepfake images as real. This could lead to unprecedented opportunities for deception and manipulation. Hence, there is an urgent need to develop deepfake detection models that are robust against adversarial attacks to ensure that detectors are able to perform equally well in both controlled and uncontrolled testing environment.

Motivated by the need for a robust deepfake detection model, this study adapts the metamorphic testing (MT) principles to help identify factors that could potentially fool deepfake detectors. In this paper, we demonstrate how an adversarial attack identified through MT can help us to evaluate the robustness of two state-of-the-art deepfake detectors, MesoInception-4 and TwoStreamNet models, and reveal their vulnerabilities.

\section{Related Work}
\textbf{Deepfake Generation: } Many techniques have been developed in the field of deepfake generation. The most prominent ones are autoencoder-decoder and GANs. Autoencoder-decoder is an unsupervised DNN that learns to reproduce given inputs\cite{ZENDRAN2021834}. Meanwhile, GAN has two neural networks - the generator and discriminator, challenging against each other to generate new synthetic instances of data that are eerily similar to the domain input \cite{GANsDescription}. It can be utilised to generate plausible new images from unlabeled original images to help with data augmentation. On the other hand, GAN can also be used unethically to generate deepfakes, making GAN a powerful yet painful invention ever in deepfake domain.

\textbf{Deepfake Detection Methods: } Deepfake detection methods are still a trending topic in the research area. Several state-of-the-art deepfake detection methods that have been proposed mostly rely on flaws in deepfake generation pipelines, i.e., visual artifacts or discrepancies \cite{ismail_elpeltagy_zaki_eldahshan_2021}.  Pairwise learning technique as discussed by the work in \cite{app10010370} processes feature extraction with common fake feature network (CFFN) in the Siamese network architecture and classification with CNN. In terms of transfer learning, bag-of-words and shallow classifiers use the bag-of-words approach to remove discriminant characteristics, which will then be fed into SVM, random forest, or Bayes classifier for binary classification \cite{app10010370}. Besides, the work by \cite{app10010370} detects deepfake images by analyzing convolutional traces and using the expectation-maximization algorithm to get local features pertaining to the convolutional generative process from GAN-based deepfake image generators.

\textbf{Deepfake Detection Models: } From the perspective of deepfake detection model creation, Afchar et al. \cite{8630761} introduces two models, Meso-4 and MesoInception-4 which are shallow networks, both with a low number of layers to focus on the mesoscopic properties of images to detect deepfake videos. In \cite{Gong2021}, the DeepfakeNet which combines both the ResNet and the Inception split-transform-merge structure, detects deepfakes based on image segmentation and deep residual network with the improved structure of fake face tempering \cite{Gong2021}. Moving on, the work in \cite{li2019exposing} performs face extraction from video frames using the dlib module. The authors applied four deep learning neural networks, namely, ResNet152, ResNet50, VGG16, and ResNet101, to uncover artifacts from face images based on resolution inconsistencies between the warped face region and its surroundings. A more generalizable deepfake detection model, TwoStreamNet developed by \cite{luo2021generalizing}, takes full advantage of image noises. By adopting multi-scale high-pass filters from the Spatial Rich Model (SRM) \cite{6197267}, the high-frequency noise and low-frequency textures are extracted and the interactions between the two modalities are formulated with their designed dual cross-modality module.

While accuracy of deepfake detection model is one of the most important metrics to measure its usefulness, it is also crucial to ensure that the models are robust and fault-tolerant in order to produce a reliable outcome under any circumstances. 
Essentially, the vulnerabilities found from a thorough evaluation and testing can be a direction for future work towards creating a more robust solution.

\textbf{Robustness Evaluation of AI Models: } To date, there are several methods to evaluate the robustness of AI models. For example, adversarial testing, robustness learning and formal verification. Tu et al. \cite{tu2021exploring} tested the robustness of multi-sensor perception systems in self-driving vehicles by introducing physically realisable and input-agnostic adversarial attacks. The experiment demonstrated that the system is vulnerable due to the easily corrupted image features. Subsequently, the findings motivate the development of the resolution such that adversarial training with feature denoising can boost the system’s robustness to such attacks noticeably. Another research by Zhang et al. \cite{10.1145/3374217} summarizes 40 types of adversarial attack that are successful in exploiting DNNs for Natural Language Processing (NLP) applications. 

\textbf{Robustness Evaluation of Deefake Detection Models: }
In terms of deepfake detectors, Trinh and Liu had investigations on the racial distribution of deepfake \cite{DeepfakeDetectorFairnessExamination}. They proved that three popular deepfake detectors display racial disparities with up to 10.7\% difference in error rate \cite{DeepfakeDetectorFairnessExamination}. As a result, the models tend to learn spurious correlations between the foreground faces and fakeness when the deepfake involves faces of different genders or races \cite{DeepfakeDetectorFairnessExamination}. This shows one of the vulnerabilities in deepfake detectors. Besides, the work by Hussain et al. showed that DNN based deepfake detectors are not robust towards adversarially modified fake videos \cite{EvaluateDeepfakeDetectorVulnerabilityWithAdversarialExamples}. These types of adversarial perturbations possess infinite risks as they can be used to attack both image and video compression codecs \cite{EvaluateDeepfakeDetectorVulnerabilityWithAdversarialExamples}. In addition, adversarial perturbations have been shown to be transferable between different models \cite{AdversarialThreatstoDeepFakeDetection}. The Universal Adversarial Perturbations technique invented further make such attacks easier and feasible as they can be easily shared among attackers \cite{AdversarialThreatstoDeepFakeDetection}.

\textbf{Our Work: } Based on the discussed literature about the importance of ensuring the robustness of deepfake detection models, this paper envisions to \textit{i) demonstrate how an adversarial example identified through metamorphic testing can degrade the performance of deepfake detection models} and \textit{ii) investigate if current state-of-the-art deepfake detection models can be bypassed when tested against a series of adversarially perturbed input images, and to what extent}. The two deepfake detection models evaluated are MesoInception-4 and TwoStreamNet. MesoInception-4 model has been widely adopted due to its efficient performance of greater than \textbf{98\%}, despite being a shallow network. Meanwhile, TwoStreamNet is a recently developed model that has the best performance in terms of cross-dataset scenarios, with an average performance of greater than \textbf{95\%} and is by far, amongst the more general deepfake detection models.
We use the FaceForensics++ dataset in our experiments, as it has the highest popularity among other deepfake datasets including Korean Deepfake Detection (KoDF), Deepfake Detection Challenge (DFDC), and Celeb-DF datasets. FaceForensics++ is commonly considered as the benchmark dataset as it involves fully unconstrained, multi-person datasets that are manipulated by four different deepfake manipulation methods including i) Deepfakes (DF), ii) Face2face (F2F), iii) Face Swap (FS) and iv) Neural Textures (NT) with different compression rates. 

\section{Methods}

This study adopts the Metamorphic Testing (MT) principle to effectively reveal adversarial examples or perturbations that influence the performance of deepfake detection models \cite{mtToAdversarialExample}. MT has been commonly used to automatically create test cases to detect software faults by utilising metamorphic relations (MR) \cite{park2021}. Every MR depicts specific properties of the testing system in terms of inputs and their expected outputs, and alleviates the test oracle problem. In cases where the correctness of the actual output is difficult to be determined, MR allows researchers to identify the causal relationship between input and output based on the expected outputs, to help infer the system's faults. In our study, we apply MT to identify adversarial examples that could potentially fool deepfake detection models. 

Based on MT, we created two sets of datasets; i) non-perturbed (original) and ii) perturbed (underwent the metamorphic transformation) datasets. Upon getting the perturbed dataset, we compare the test results of the two models on the non-perturbed and perturbed datasets as illustrated in Figure \ref{fig:metamorphicTestingFlow}. If the models' performances on the non-perturbed and perturbed datasets differ from each other, the perturbation factor is identified as a potential perturbation that changes the behaviour of the model.

\begin{figure}[h!]
    \centering
    \includegraphics[width=0.3\textwidth]{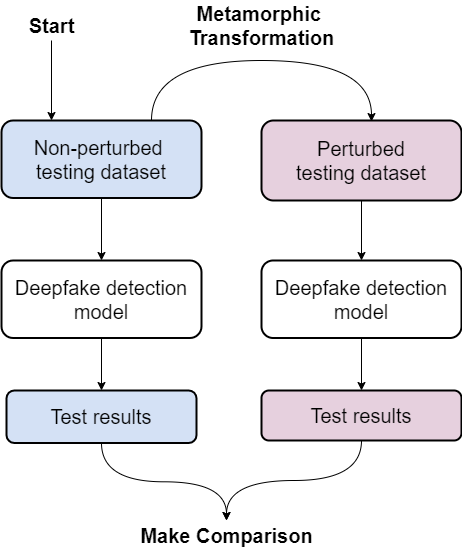}
    \caption{Metamorphic testing flow chart.}
    \label{fig:metamorphicTestingFlow}
\end{figure}

In this study, we deem makeup as a possible perturbation factor (or noise) since makeup is common in publicly available images and to date it is not identified by any source as a deepfake factor. As discussed by Tolosana et al., the four main types of face manipulation techniques established are i) entire face synthesis, ii) identity swap, iii) attribute manipulation, and iv) expression swap \cite{surveyOnDeepfakes}. Specifically, the techniques include creation of entire non-existent face images, face swapping between people, face editing or retouching such as alteration of the age and addition of synthetic objects onto the faces as well as modification of facial expression \cite{surveyOnDeepfakes}. 

The flow of our makeup application is as shown in Figure \ref {fig:makeupFlow}. Dlib library is utilised to identify the 68 facial landmarks of the targeted face. The positions of the facial landmarks of the intended regions as shown in Figure \ref {fig:makeupFlow} 
are then obtained to facilitate the interpolation of the boundaries of the makeup region. After retrieving all the coordinates within the makeup region, RGB-colored polygons are drawn at the coordinates accordingly via the cv2.fillPoly() or cv2.fillConvexPoly() methods of the OpenCV libraries. We also utilised the Gaussian Blur filter on the makeup components eyeshadow, blushes and lipstick so that the colored polygons blend in better with the image and appear more naturally in order to not fool the human eye.

\begin{figure}[h!]
    \centering
    \includegraphics[width=0.45\textwidth]{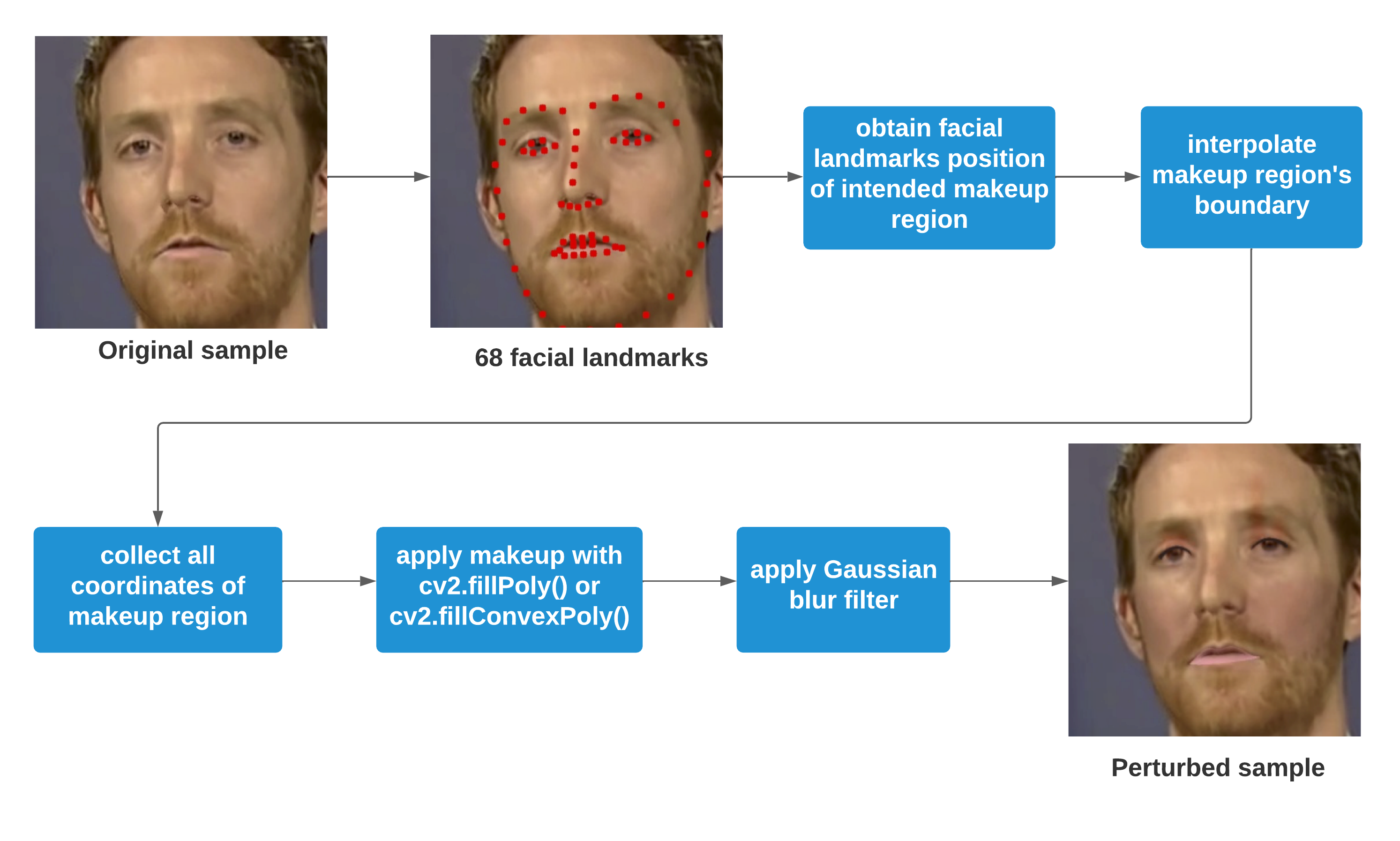}
    \caption{Makeup application flow chart.}
    \label{fig:makeupFlow}
\end{figure}

In the context of this study, the MR is defined such that the perturbed testing dataset,  $i'$, is synthesized by adding noise on the testing dataset, $i$. The deepfake detection result is expected to remain unchanged when tested with both $i$ and $i'$. Then, given the set of makeup artifacts $M$, the coordinates of the facial landmarks obtained, $F$, and the corresponding indices that match the makeup artifact with its targeted facial landmark, $X$, the perturbed image can be represented by equation \ref{eq:mt} below:
\begin{equation} \label{eq:mt}
    i' = {f^x(i, m^x), f \in \mathbb{F}\ \mathrm{and}\ m \in \mathbb{M}}
\end{equation}
where a test image, $i$, is applied with makeup artifact, $m^x$, at the facial region identified by facial landmarks, $f^x$, iteratively for all $x$ to produce the final perturbed image, $i'$. In our case, we consider makeup artifacts as the combination of eyeliner, eyeshadow, blushes, and lipstick. Based on the results, since the models' performances on the non-perturbed and perturbed datasets differ from each other, makeup is identified as a potential perturbation that changes the behavior of the models. The perturbation, therefore, acts as our adversarial example in this study. The experiment details and test results are discussed in the Experiments section below.

\section{Experiments}
\subsection{Datasets}
We used the FaceForensics++ video dataset with a compression rate of 23 (HQ) for all our experiments as this is the common compression rate used for the experiments in the original MesoNet and TwoStreamNet papers. For the training and validation datasets, we used the forged videos using the Face2face method and their originals. To evaluate the models' performance on the cross-dataset scenarios, aside from the original videos and F2F forged videos, we also included the forged videos using DF, FS, and NT methods for the testing datasets. Both non-perturbed and perturbed versions of the testing datasets are taken into account to gauge the models’ robustness against the makeup perturbation applied. 

We conducted two different experiments to simulate both MesoInception-4 and TwoStreamNet's dataset setup- i) \textit{Sub-sample}: This is based on the MesoNet's paper where only around 440 videos are used for both training and testing out of the 1000 videos \cite{8630761} ii) \textit{Complete}: This is based on the TwoStreamNet's paper setting, where all 1000 videos are used \cite{luo2021generalizing}. One of the main strengths of the MesoInception-4 model is its capability to detect deepfake efficiently while using a small dataset and the least amount of time for training. Thus, we also perform testing of TwoStreamNet on the sub-sample dataset to examine if the model can achieve consistent performance under such constraint.

For both experiments, we performed the same preprocessing steps. First, we extracted face images of size $256 \times 256$ from the video datasets and aligned them using the pipeline module provided by the authors of MesoNet \cite{8630761}. For DF, FS, and NT datasets, only test partitions are used for cross-dataset evaluation. After applying the makeup artifacts on the test image datasets to produce the perturbed image datasets \footnote{The code and perturbed datasets will be shared on our GitHub repository upon publication}, images that failed to be applied with makeup are discarded and the numbers of remaining face images in each subset of the perturbed and non-perturbed testing datasets are balanced before performing the test. The complete test images dataset are also extracted in such a way that the number of images is the same as the sub-sample dataset's test images for consistent comparison. The details of both the sub-sample and complete datasets' dimensions are detailed in Table \ref{table:subsampleDataset} and Table \ref{table:completeDataset}.
\begin{table}[h!]
\caption{\textbf{Sub-sample} dataset dimensions.}
     \centering
     \begin{tabular}{|l|l|}
        \hline
        Dataset        & Number of images \\ \hline
        Training set   & 4050 real, 4050 deepfake (F2F) \\ \hline
        Validation Set & 450 real, 450 deepfake (F2F) \\ \hline
       \multirow{2}{*}{Testing Set} &     \begin{tabular}[c]{@{}l@{}}
            Non-perturbed: \\ 
            2639 real,  10556 deepfake \\
            (2639 F2F + 2639 DF + 2639 FS + 2639 NT)
        \end{tabular} \\ \cline{2-2} &
        \begin{tabular}[c]{@{}l@{}}
            Perturbed:\\ 
            2639 real, 10556 deepfake \\
            (2639 F2F + 2639 DF + 2639 FS + 2639 NT)
        \end{tabular} \\ \hline              
    \end{tabular}
\label{table:subsampleDataset}
\end{table}
\begin{table}[h!]
\caption{\textbf{Complete} dataset dimensions.}
     \centering
     \begin{tabular}{|l|l|}
        \hline
        Dataset        & Number of images \\ \hline
        Training set   &  366714 real, 366837 deepfake (F2F) \\ \hline
        Validation Set & 3454 real, 3454 deepfake (F2F) \\ \hline
       \multirow{2}{*}{Testing Set} &     \begin{tabular}[c]{@{}l@{}}
            Non-perturbed: \\ 
            2639 real,  10556 deepfake \\
            (2639 F2F + 2639 DF + 2639 FS + 2639 NT)
        \end{tabular} \\ \cline{2-2} &
        \begin{tabular}[c]{@{}l@{}}
            Perturbed:\\ 
            2639 real, 10556 deepfake \\
            (2639 F2F + 2639 DF + 2639 FS + 2639 NT)
        \end{tabular} \\ \hline              
    \end{tabular}
\label{table:completeDataset}
\end{table}

\subsection{Evaluation Metric}
The evaluation of the two deepfake detection models is done by comparing the three metrics, accuracy, specificity, and recall. We define our confusion matrix as shown in Table \ref{table:confusionMatrix} below:
\begin{table}[h!]
    \caption{Confusion matrix.}
    \centering
    \begin{tabular}{|ll|ll|}
    \hline
    \multicolumn{2}{|l|}{\multirow{2}{*}{}} & 
        \multicolumn{2}{l|}{Predictions} \\ \cline{3-4} 
            \multicolumn{2}{|l|}{}  & 
                \multicolumn{1}{l|}{Deepfake} & Real \\ \hline
                \multicolumn{1}{|l|}{\multirow{2}{*}{Actual}} & Deepfake & \multicolumn{1}{l|}{True Positive (TP)}  & False Negative (FN) \\ \cline{2-4} 
                \multicolumn{1}{|l|}{} & Real  &
                \multicolumn{1}{l|}{False Positive (FP)} & True Negative (TN)  \\ \hline
    \end{tabular}
    \label{table:confusionMatrix}
\end{table}
From the confusion matrix, we then calculate the accuracy, specificity, and recall as shown in equations \ref{eq:accuracy}, \ref{eq:recall} and \ref{eq:specificity}:
\begin{equation} \label{eq:accuracy}
    Accuracy(\%) = {\frac{TP + TN}{TP + FN + FP + TN} \times 100}
\end{equation}
\begin{equation} \label{eq:recall}
    Recall(\%) = {\frac{TP}{TP + FN} \times 100}
\end{equation}
\begin{equation} \label{eq:specificity}
    Specificity(\%) = {\frac{TN}{FP + TN} \times 100}
\end{equation}
Accuracy is chosen as it is the common evaluation metric used for deepfake detector. It informs us on the degree of correctness and conformity of models' performance to the truth values. However, it suffers from the accuracy paradox, where a high accuracy model may fail to capture essential information in the classification task \cite{accuracy_paradox}. 
Hence, we include specificity and recall as FN and FP are equally important to gain more insights into the models' performance. A high recall indicates that the model has done well at identifying TP, while a low recall indicates high FN. Thus, recall is well suited in output-sensitive context such as predicting deepfake, or predicting cancer. Meanwhile, specificity value shows how good a model is at avoiding false alarms. High specificity is well suited for areas where the focus is on TP and FP, such as recommendation engines. 

\subsection{Implementation}
Model training wise, the pre-trained weights for the MesoInception-4 model is provided by the original authors at the GitHub repository \cite {8630761}. In contrast, there is a problem of replicability with TwoStreamNet as the author did not disclose certain parameters' values as well as the final weights due to confidentiality issues (after we contacted the authors). Hence, we had retrained the TwoStreamNet model and determined the parameter settings empirically. The training dataset for the two TwoStreamNet model's settings are as shown in Table \ref{table:subsampleDataset} and Table \ref {table:completeDataset} respectively. Training of the TwoStreamNet model is done by first initializing the model using the pre-trained weights of ImageNet provided by the author \cite{luo2021generalizing}. Then,  we trained the model with the parameters of a batch size of 16 and an Adam optimiser with a learning rate of 0.0002. 

The sub-sample and complete experimental datasets each contain genuine and four categories of deepfake manipulation methods' non-perturbed and perturbed datasets. The four categories are i) F2F, ii) DF, iii) NT, and iv) FS. First, the non-perturbed datasets of all categories are fed into the models for evaluation. Upon getting the test outputs, we computed their evaluation metrics which are accuracy, recall, and specificity accordingly. The test results collected at this stage serve as the benchmark which we can then compare with at the following stage. Then, we performed the models' evaluation in a similar manner for the other four deepfake manipulation methods' dataset.

In the analysis stage, the evaluation metrics of the perturbed datasets of each category are compared with their respective benchmarking test results to verify the defined MR. This allows us to gauge the models' robustness by examining if their performance changes with the makeup adversarial example. By applying MT principles, our assumption defined in our MR is that the classification results shall not change after the input datasets are perturbed. We also discuss the model's performance in cross-datasets setting to evaluate its generalisation capability, where the model is trained on one dataset and tested on another dataset.


For TwoStreamNet model, we conducted an additional comparison of its test results. By comparing its performance on the sub-sample and complete datasets (Tables \ref{table:subsampleDataset} and \ref{table:completeDataset}), we could determine if the training dataset size does influence the deep learning model's robustness against the perturbed dataset.

\subsection{Results and Discussion}
As shown in Table \ref{table:modelAccuracy}, TwoStreamNet model's performance in this study does not resemble a consistent result as published by the authors, in which an accuracy of \textbf{$>$99\%} is achieved across datasets for all 4 manipulation methods \cite{luo2021generalizing}. We found that this might be due to the different training parameters settings and hardware used for experiments. 

Comparing the performance of both MesoInception-4 and TwoStreamNet models on the non-perturbed and perturbed sub-sample testing datasets as visualised in Figure \ref {table:modelAccuracy}, both models exhibit a similar pattern, which is a drop in their performance when tested with the perturbed datasets. The degrade in performance is most drastic in within-dataset testing, and less in cross-dataset testing. This is mainly due to the already poor performance in cross-dataset testing.
\begin{table}[h!]
\caption{Test \textbf{accuracy} results of MesoInception-4 and TwoStreamNet (trained on F2F) on \textbf{sub-sample} dataset (\%).}
    \centering
    \begin{tabular}{|l|l|l|l|}
    \hline
    Model & Dataset & Non-perturbed & Perturbed \\ \hline
    \multirow{4}{*}{MesoInception-4} 
        & F2F & \textbf{86.00} & 52.67 \\ \cline{2-4} 
        & DF & 52.24 & 50.13 \\ \cline{2-4} 
        & FS & 51.02 & 50.09 \\ \cline{2-4} 
        & NT & 49.22 & 50.09 \\ 
        \hline
    \multirow{4}{*}{TwoStreamNet} 
        & F2F & \textbf{99.62} & 55.48 \\ \cline{2-4} 
        & DF & 55.95 & 50.04  \\ \cline{2-4} 
        & FS & 50.68 & 49.98  \\ \cline{2-4} 
        & NT & 50.97 & 50.11  \\ 
        \hline
    \end{tabular}
\label{table:modelAccuracy}
\end{table}

By examining the recall and specificity values calculated in Table \ref{table:modelRecall} and Table \ref{table:modelSpecificity}, perturbation with makeup application had caused the recall value to decrease and the specificity value to increase across all four deepfake manipulation methods. This indicates an increase in the number of predicted real images after makeup  application. \textbf{The makeup that has been added with the intention to fool the model has indeed successfully confused both models into identifying most images as genuine, regardless of their actual state.} This observation is consistent across all the four deepfake manipulation methods' testing datasets, including F2F, NT, DF, and FS. These comparisons show that the two deepfake detection models, MesoInception-4 and TwoStreamNet, are not robust towards makeup. Both models can be easily fooled to falsely identify deepfakes as genuine when the input images are perturbed with makeup; a rather simple, yet powerful adversarial attack.

\begin{table}[h!]
\caption{Test \textbf{recall} results of MesoInception-4 and TwoStreamNet (trained on F2F) on \textbf{sub-sample} dataset (\%).}
    \centering
    \begin{tabular}{|l|l|l|l|}
    \hline
    Model & Dataset & Non-perturbed & Perturbed \\ \hline
    \multirow{4}{*}{MesoInception-4} 
        & F2F & 74.31 & 5.42 \\ \cline{2-4} 
        & DF & 6.78 & 0.34 \\ \cline{2-4} 
        & FS & 4.36 & 0.27 \\ \cline{2-4} 
        & NT & 0.76 & 0.27 \\ 
        \hline
    \multirow{4}{*}{TwoStreamNet} 
        & F2F & 99.51 & 11.03 \\ \cline{2-4} 
        & DF & 12.16 & 0.15 \\ \cline{2-4} 
        & FS & 1.63 & 0.04 \\ \cline{2-4} 
        & NT & 2.20 & 0.30 \\ 
        \hline
    \end{tabular}
\label{table:modelRecall}
\end{table}
\begin{table}[h!]
\caption{Test \textbf{specificity} results of MesoInception-4 and TwoStreamNet (trained on F2F) on \textbf{sub-sample} dataset (\%).}
    \centering
    \begin{tabular}{|l|l|l|l|}
    \hline
    Model & Dataset & Non-perturbed & Perturbed \\ \hline
    \multirow{4}{*}{MesoInception-4} 
        & F2F & 97.69 & 99.92 \\ \cline{2-4} 
        & DF & 97.69 & 99.92  \\ \cline{2-4} 
        & FS & 97.69 & 99.92  \\ \cline{2-4} 
        & NT & 97.69 & 99.92  \\ 
        \hline
    \multirow{4}{*}{TwoStreamNet} 
        & F2F & 99.73 & 99.92 \\ \cline{2-4} 
        & DF & 99.73 & 99.92  \\ \cline{2-4} 
        & FS & 99.73 & 99.92  \\ \cline{2-4} 
        & NT & 99.73 & 99.92  \\ 
        \hline
    \end{tabular}
\label{table:modelSpecificity}
\end{table}

Following this, based on Table \ref{table:modelAccuracy}, we also observed that \textbf{both the MesoInception-4 and TwoStreamNet models only achieve good performance on the non-perturbed F2F sub-sample testing dataset on which they are originally trained.} Their accuracy degrades significantly with an approximate value of \textbf{~40\%} when being tested on unseen domains (datasets which are not used for training). From these results, we can also deduce that both MesoInception-4 and TwoStreamNet are not robust enough in dealing with cross-datasets scenarios. Most of the time, machine learning models work well on seen data but in a high-dimensional space, a small perturbation on the image pixel which is not usually visible to the human eye, may cause dramatic change in the dot product of the neural network. Our study simply proves that indeed state-of-the-art deepfake detection models are susceptible to adversarial attacks and can have huge implications to society.

Using the complete dataset, 
the TwoStreamNet model's performance on non-perturbed testing datasets across all four deepfake manipulation methods is found to be similar to the results when it is trained on the sub-sample dataset, with up to a minimal of +/- \textbf{0.11\%} difference as shown in Table \ref{table:modelAccuracyAll}. However, when comparing the accuracy on the perturbed testing datasets, we can see an obvious improvement in the accuracy of TwoStreamNet on the perturbed F2F testing dataset, from \textbf{55.48\%} to \textbf{74.18\%}, while the other three deepfake testing datasets' performance remains similarly low at around \textbf{50\%} compared to the sub-sample setting results. This is due to \textbf{the ability of the TwoStreamNet model to correctly identify more actual F2F deepfake images as deepfake when trained on a larger dataset} as revealed by the drastic increase in its percentage of recall, as shown in Table \ref{table:modelRecallAll}. On the other hand, the specificity values of all 4 deepfake manipulation methods' testing datasets on both dataset size settings do not differ much. This shows that the model does not improve its ability to identify real images as real ones even though it is trained on a larger dataset. The test results also further support the earlier results on the sub-sample setting where the TwoStreamNet model performs the best in the F2F dataset that it is trained on. Overall, these findings allude to the common understanding that changes in the size and quality of training data impacts the performance of deep learning models.

\begin{table}[h!]
\caption{Test \textbf{accuracy} results of TwoStreamNet model (trained on F2F) on \textbf{complete} dataset (\%).}
    \centering
    \begin{tabular}{|l|l|l|l|}
    \hline
    Model & Dataset & Non-perturbed & Perturbed \\ \hline
    \multirow{4}{*}{TwoStreamNet} 
        & F2F & 98.52 & 74.18 \\ \cline{2-4} 
        & DF & 52.88 & 50.00  \\ \cline{2-4} 
        & FS & 50.47 & 50.02  \\ \cline{2-4} 
        & NT & 50.21 & 50.00  \\ 
        \hline
    \end{tabular}
\label{table:modelAccuracyAll}
\end{table}
\begin{table}[h!]
\caption{Test \textbf{recall} results of TwoStreamNet model (trained on F2F) on \textbf{complete} dataset (\%).}
    \centering
    \begin{tabular}{|l|l|l|l|}
    \hline
    Model & Dataset & Non-perturbed & Perturbed \\ \hline
    \multirow{4}{*}{TwoStreamNet} 
        & F2F & 97.50 & 48.35 \\ \cline{2-4} 
        & DF & 6.21 & 0.00 \\ \cline{2-4} 
        & FS & 1.40 & 0.04 \\ \cline{2-4} 
        & NT & 0.87 & 0.00 \\ 
        \hline
    \end{tabular}
\label{table:modelRecallAll}
\end{table}
\begin{table}[h!]
\caption{Test \textbf{specificity} results of TwoStreamNet model (trained on F2F) on \textbf{complete} dataset (\%).}
    \centering
    \begin{tabular}{|l|l|l|l|}
    \hline
    Model & Dataset & Non-perturbed & Perturbed \\ \hline
    \multirow{4}{*}{TwoStreamNet} 
        & F2F & 99.55 & 100 \\ \cline{2-4} 
        & DF & 99.55 & 100  \\ \cline{2-4} 
        & FS & 99.55 & 100  \\ \cline{2-4} 
        & NT & 99.55 & 100  \\ 
        \hline
    \end{tabular}
\label{table:modelSpecificityAll}
\end{table}

\begin{table}[h!]
\caption{Test \textbf{accuracy} results of TwoStreamNet model (trained on F2F) on \textbf{complete} dataset (\%).}
    \centering
    \begin{tabular}{|l|l|l|l|}
    \hline
    Model & Dataset & Non-perturbed & Perturbed \\ \hline
    \multirow{4}{*}{TwoStreamNet} 
        & F2F & 98.52 & 74.18 \\ \cline{2-4} 
        & DF & 52.88 & 50.00  \\ \cline{2-4} 
        & FS & 50.47 & 50.02  \\ \cline{2-4} 
        & NT & 50.21 & 50.00  \\ 
        \hline
    \end{tabular}
\label{table:modelAccuracyAll}
\end{table}

In the last stage, we manually extracted a total of 505 deepfake images from the four deepfake manipulation methods' testing datasets where the subjects already have makeup on originally. The test results shown in Table \ref{table:modelOnMakeupFakeDataset} reflect low recall values with more false negatives compared to true positives. A similar trend to earlier experiments when the makeup is applied intentionally via our makeup application is observed. Hence, we can treat our makeup application as simply augmenting the original makeup images. This further verifies our hypothesis that \textbf{natural makeup or makeup via our application can lead to both the models falsely detect the deepfake images as genuine}.

\begin{table}[h!]
\caption{Test results of models on \textbf{deepfake} dataset with \textbf{natural makeup}.}
    \centering
    \begin{tabular}{|l|ll|}
    \hline
    Model  & \multicolumn{2}{l|}{Results}  \\ \hline
    \multirow{3}{*}{TwoStreamNet}    
        & \multicolumn{1}{l|}{TP}     & 241     \\ \cline{2-3} 
        & \multicolumn{1}{l|}{FN}     & 264     \\ \cline{2-3} 
        & \multicolumn{1}{l|}{Recall(\%)} & 47.72 \\ \hline
    \multirow{3}{*}{MesoInception-4} 
        & \multicolumn{1}{l|}{TP}     & 161     \\ \cline{2-3} 
        & \multicolumn{1}{l|}{FN}     & 344     \\ \cline{2-3} 
        & \multicolumn{1}{l|}{Recall(\%)} & 31.88 \\ \hline
    \end{tabular}
\label{table:modelOnMakeupFakeDataset}
\end{table}

\section{Conclusion and Future Work}
In this work, we present a method to evaluate the robustness of state-of-the-art deepfake detection models through metamorphic testing. 
Using makeup applications as adversarial examples, we found that it is capable of fooling both the MesoInception-4 and TwoStreamNet models into falsely identifying deepfakes as real images. Both models consistently fail to detect deepfake images on datasets obtained from the four deepfake manipulation methods, simply by adding makeup to the input images. Our experimental findings prove that \textit{i) MT has the potential to help identify factors that can serve as adversarial attacks to create intended vulnerabilities in deepfake detection models,} and \textit{ii) existing deepfake detection models are not robust against adversarial attacks and cross-dataset scenarios}, and their performances are highly influenced by the training dataset size. Our preliminary findings show that a larger training dataset size does not only improves the performance of detectors, but also increases its robustness towards makeup as an adversarial example. We believe that this study can help spur new research opportunities and directions in the field of deepfake detection, specifically toward devising robust detectors and dealing with adversarial attacks. Most importantly, we have shown how easy it is to fool sophisticated deep learning-based models such as MesoInception-4 and TwoStreamNet, simply by adding intentional yet subtle noise into the input images. 
\section*{Acknowledgments}
We would like to thank the authors of TwoStreamNet \cite{luo2021generalizing} for responding to our queries. This work was partly supported by the Advanced Engineering Platform's Cluster Funding (AEP-2021-Cluster-04), Monash University, Malaysia.

\printbibliography 
\end{document}